\documentclass[runningheads]{llncs}
\usepackage[T1]{fontenc}
\usepackage{graphicx}
\graphicspath{{figures/}}
\usepackage{tikz-cd}
\usepackage{amsmath}
\usepackage{tabularray}
\usepackage{listings}
\begin{document}
\title{Deep probabilistic logic programming for diagnostic reasoning from incomplete information: \\
  A case study in stroke detection}
\titlerunning{Deep probabilistic logic programming for diagnostic reasoning}
%
\author{Felix Weitkämper \orcidID{0000-0002-3895-8279} \and Monchito Avila \and \\
  Elizabeth Nanjala \and Siska \and Grace Zawadi}
\authorrunning{F. Weitkämper et al.}
%
\institute{German University of Digital Science\\
\email{felix.weitkaemper@german-uds.de}}
\maketitle              
\begin{abstract}
  In medical applications, raw data is frequently associated with significant privacy concerns, lending particular importance to the encoding of summary statistics from the literature.
  On the other hand, deep learning has become an invaluable tool for assessing symptoms based on visual or auditory sensor data.
    DeepProbLog allows for an extensible neuro-symbolic approach that accommodates connectionist components to analyse patient images within a transparent and rigorous probabilistic framework, namely probabilistic logic programming under the distribution semantics. 
    Framed as a case study in stroke detection from multimodal data, this contribution explores the pathway from summary statistics available in the literature to a DeepProbLog-based diagnostic system.
    It suggests a workflow using established maximum entropy techniques to complete available probabilistic information and the probabilistic logic programming system ProbLog 2 to move from the entropy-maximising causal model to a discriminative neuro-symbolic model expressible within DeepProbLog.
    The relative performance of models derived from less complete data is analysed alongside the potential of the probabilistic inductive logic programming system ProbFOIL 2 for compressing large discriminative models, and the perspectives and implications of using DeepProbLog for diagnostic reasoning are discussed. 

\end{abstract}
\section{Introduction}
Privacy considerations are a defining feature of medical applications and the circumstances in which they are developed and operate.
In particular, raw medical data should be used as sparingly as possible, and its use is often strictly regulated.
Hence, day-to-day decisions are made on the basis of published summary statistics and other information taken from the medical literature.
Those include the likelihood of specific symptoms and symptom combinations, as well as general epidemiological survey data on the likelihoods of illnesses and symptoms in the general population.

Hence, when designing a decision support system for medical diagnosis, one should be able to incorporate such incompletely specified probabilistic information.
On the other hand, deep learning has made huge progress in determining symptoms detectable by visual or auditory data, such as detecting facial asymmetry or dysarthric speech \cite{ElhanashiSZ24,BanerjeeSJGRRSDG25,DuaS25}.
As a neuro-symbolic architecture explicitly designed to combine a stringent modular probabilistic framework in a well-developed ecosystem (probabilistic logic programming in ProbLog 2) with connectionist components for processing sensor data (neural predicates), DeepProbLog \cite{ManhaeveDKDD21} seems particularly well-suited to this category of tasks.


In our case study, the goal is to predict the probability of suffering a stroke or transient ischemic attack (TIA) 
when provided with information on a prospective  patient's current symptoms.
Most of the information on an individual's condition is  obtained from the individual (or another user in the individual's vicinity) directly;
for instance,  information is to be provided on whether they are experiencing weakness in arms or legs, whether their vision is impaired or whether their speech is impacted.
Additionally, facial asymmetry is to be assessed using a neural computer vision model.

To evaluate the likelihood of a stroke or TIA given the user's input, we use statistical information available from the literature.
While in principle it would be possible to infer probabilities directly from data, raw medical data is subject to significant privacy concerns.
Hence, using public information available in the literature reduces exposure of potentially sensitive information,
and also avoids significant regulatory obstacles to the initial development and deployment of the system.

We use DeepProbLog to integrate the neural and symbolic information available to the system and the summary statistics on the symptomatics and general prevalence of stroke reported in the literature.
As DeepProbLog assumes neural predicates as facts rather than as (soft) evidence, DeepProbLog programs are constrained to a representation in which the target variable depends on the input variables, one of which is neural (a \textit{diagnostic} model).
While the stroke recognition literature has extensively characterised the sensitivity of symptom-based screening tools such as FAST and  its extensions \cite{ClausBBLRIW24,ChenZXGYZLWL21}, the positive predictive value of specific symptom combinations, that is, the probability of stroke given a particular co-occurrence pattern of symptoms, has not been its focus.

Hence, we proceed with a two-step approach, in which we first create a generative probabilistic (ProbLog) model of stroke and its symptoms which follows the causal order, and then use ProbLog conditional inference to deduce the diagnostic model required by the DeepProbLog architecture.     
\section{Encoding probabilistic knowledge}

In our setting, probabilistic information is provided as conditional probabilities.
Those can connect a disease and its symptoms, or they could connect risk factors to a disease.
Usually, probabilistic information is partial, in that it does not allow for the reconstruction of the full probability distribution  across all possible combinations of illness and symptoms.
Hence, we need to follow some guideline on which probability distribution to adopt in the presence of partial probabilistic information.
Various arguments have been advanced in favour of choosing, among  compatible distributions, the one which \emph{maximises entropy} \cite{ShoreJ80,ParisV90,ParisV97,Williamson04}.

\begin{definition}
  Let $\sigma$ be a set of Boolean random variables (interpreted as propositions) and let $\mathcal{P}$ be a finite set of equations of the form $\mu(\varphi \mid \psi) = \theta$, where $\varphi$ and $\psi$ are propositional formulas.
  Then a probability distribution $\mu$ on $\sigma$ is \emph{compatible} with $\mathcal{P}$ if it satisfies all equalities in $\mathcal{P}$.
  The \emph{entropy} of $\mu$ is defined as
  \[-\sum_{\omega\in 2^\sigma}\mu(\omega) \log_2\mu(\omega),\] where $0\log_2(0)$ is 0, and $\mu$ \emph{maximises entropy for $\mathcal{P}$} if no other probability distributions on $\sigma$ compatible with $\mathcal{P}$ has greater entropy than $\mu$.    
\end{definition}

While maximising entropy analytically can be difficult in principle, Williamson \cite[Theorem 5.1]{Williamson04} gives a very useful criterion for conditional independence in the maximum entropy distribution.
\begin{proposition}\label{prop:maxent-ind}
  Let the \emph{constraint graph} of a set of constraints $\mathcal{P}$ be the undirected graph on $\sigma$ that has an edge between $A,B\in \sigma$ whenever $A$ and $B$ co-occur in a constraint of $\mathcal{P}$.
  Then the following holds: Let $X,Y,Z \in \sigma$.
  Then if $Z$ separates $X$ from $Y$ in the constraint graph of $\mathcal{P}$, then $X$ is conditionally independent of $Y$ given $Z$ in any probability distribution $\mu$ that maximises entropy for $\mathcal{P}$.
\end{proposition}
Thus, if all of our knowledge is of the form $\mu(\mathrm{symptom}\mid \mathrm{illness}) = p$ for some $p$, Proposition \ref{prop:maxent-ind} implies that all symptoms will be  conditionally independent given the illness in the entropy-maximising distribution.

On closer inspection, medical knowledge is rarely in the form of plain conditional probabilities.
For instance, in an analysis of stroke symptoms, what is reported is not usually the raw conditional probability of a symptom given a stroke, but rather some clinical effort is undertaken to only record those symptoms for which the stroke is believed to be causal;
symptoms known to be caused by other, prior health conditions are disregarded.
We argue that to capture this, we need to prescribe more structure to the entropy maximisation problem.
In our application scenario, we assume that for any symptom $S$, we have an equality $S = S_{\mathrm{stroke}} \lor S_{\textrm{other causes}}$, where we treat the literature reports on $S_{\mathrm{stroke}}$ and  $S_{\textrm{other causes}}$ separately.
The conditional probability of $S_{\mathrm{stroke}}$ given neither stroke nor TIA is set as 0.
Note that standard entropy calculations (the entropy chain rule) show that adding this deterministic constraint does not impact the maximum entropy distributions of  $S_{\mathrm{stroke}}$ and  $S_{\textrm{other causes}}$.

For our present case study, we restrict ourselves to information from a single study with a fairly large number of participants \cite{ClausBBLRIW24}, as aggregating probabilistic information inferred from different study settings poses additional methodological challenges.
Their study included 900 patients with first-ever stroke or TIA from 2002--2016, as well as a survey with 3,854 other respondents asking for the occurrence of symptoms otherwise associated with stroke within a 2-month window, both part of the population-based Rotterdam Study \cite{Hofman15}. The prevalence and co-occurrence of focal neurological symptoms were determined for each severity group (TIA, minor stroke, and major stroke).
To assess the sufficiency of different levels of modelling complexity for this use case, we model both the relative occurrence of all specific symptom combinations from \cite{ClausBBLRIW24} as their Figure 2 and the summary information only on isolated symptoms reported in their Table 2.
The figure and the table are presented as Figure \ref{fig:full_symptoms} and Table \ref{tab:simplified_symptoms} below. 
\begin{figure}
  \includegraphics[width=\textwidth]{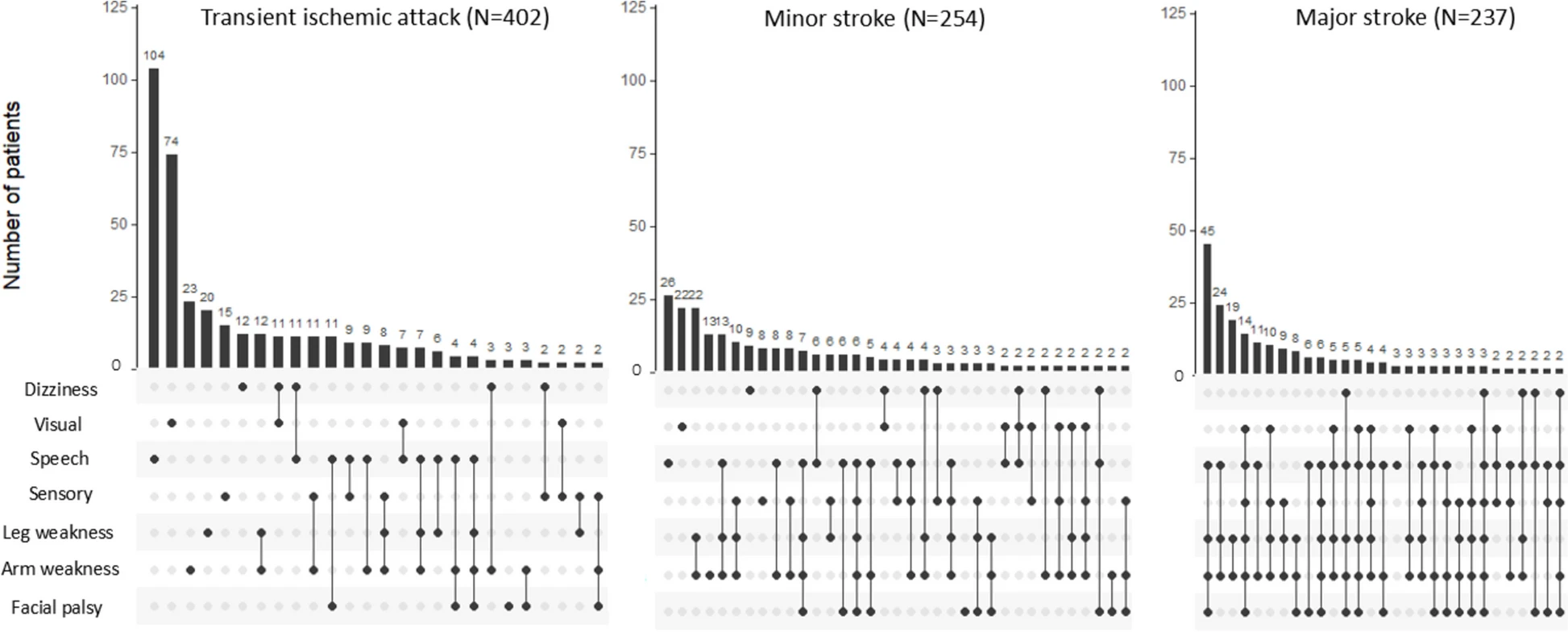}
  \caption{Concurrence of neurological symptoms by TIA and stroke severity. Legend: These intersection diagrams show the patterns of co-occurrence of different symptoms. Rows represent the types of symptom and columns represent their combinations. All symptoms that are part of a given combination are shown as black dots connected by a vertical black line. A single dot without a line implies the symptom occurred in isolation. The number of participants with a given combination of symptoms is shown as a vertical bar on top of the matrix. A minimum of two participants per combination is shown, single participant combinations are not shown. Figure, caption and legend reproduced from Claus et al.\ \cite[Figure 2]{ClausBBLRIW24}. CC-BY 4.0.}
  \label{fig:full_symptoms}
\end{figure}

\begin{table}
\centering
\begin{tblr}{
    colspec = {lllll},
    row{1} = {font=\bfseries}
  }
                          & TIA        & Minor stroke  & Major stroke  & Respondents \\ 
Symptom                   & N~= 409    & N~= 254      & N~= 237      & N~= 3,854            \\ \hline
Facial               & 36 (8.8)   & 47 (18.5)    & 126 (53.2)   & 39\textsuperscript{a} (1.0)             \\
Arm               & 96 (23.5)  & 126 (49.6)   & 211 (89.0)   & 303\textsuperscript{b} (7.8)            \\
Leg                & 69 (16.9)  & 89 (35.0)    & 183 (77.2)   & 303\textsuperscript{b} (7.8)            \\
Speech              & 176 (43.0) & 99 (39.0)    & 167 (70.5)   & 39\textsuperscript{a} (1.0)             \\
Sensory         & 58 (14.2)  & 68 (26.8)    & 80 (33.8)    & 243 (6.3)            \\
Visual             & 100 (24.4) & 46 (18.1)    & 62 (26.2)    & 181 (4.7)            \\
\end{tblr}
\caption{Table of symptom combinations reported by Claus et al.\ \cite[Table 2]{ClausBBLRIW24}. Slightly abridged by removing rows for which respondent data is unavailable. ``Facial'' refers to facial palsy, ``Arm'' and ``Leg'' refer to arm and leg weakness respectively, ``Speech'' to speech or language impairment and ``Visual'' to visual symptoms. The two symptoms each marked by (a) and (b) respectively have been put to respondents as a single question; the table notes the joint figure.}
\label{tab:simplified_symptoms}
\end{table}

\section{The generative model}
\label{sec:generative}

As a representation language for the probability distributions we obtain from entropy maximisation, we use the probabilistic logic programming language ProbLog and its implementation ProbLog 2 \cite{FierensVanDenBRSGTJD2015}.
An in-depth treatment of the language can be found in the overview \cite{DeRaedtK15}, and we just briefly rehearse the main concepts, restricting ourselves to the case of a propositional signature $\sigma$. 

ProbLog represents probabilistic dependencies through \emph{probabilistic rules} of the form
\[ p :: H \leftarrow B_1, \dots, B_n.\]
where $p\in (0,1]$ is the probability associated with the rule, $H$ is an atom called its \emph{head} and $B_1, \dots, B_n$ are literals collectively called its \emph{body}.
If the body is empty, this is known as a \emph{probabilistic fact}, written $p :: H.$   

The probabilistic semantics of a ProbLog program decomposes every such probabilistic rule into an ordinary Datalog clause
\[H \leftarrow B_1, \dots, B_n, E_r\]
and a probabilistic fact $p :: E_r$, where $E_r$ is a new fact unique to this particular rule (known as an \emph{error term}).

This induces a probability distribution on $\sigma$ in two steps.
First, all probabilistic facts and error terms $E_r$ are treated as \emph{independent} Boolean random variables with their annotated probability.
This yields a probability distribution on the set of error terms and propositions involved in probabilistic facts.
Then, for every truth value assignment on the error terms, those Datalog clauses whose associated error term is true form a Datalog program in the signature $\sigma$.
If this program is acyclic in the sense that the dependency graph with arrows from every body atom of a rule to its head atom is acyclic, iteratively evaluating those clauses and setting head atoms to be true whenever their body atoms are true will yield a truth value assignment on $\sigma$.
Hence, the  probability distribution on the error induces a corresponding probability distribution on $\sigma$ itself.
The ProbLog 2 system allows for querying this distribution for the conditional probability of a given proposition given some evidence. 

\begin{example}
  Consider the program with probabilistic logic program rule
  \[0.2 :: \mathrm{weakness} \leftarrow \mathrm{stroke}.\]
  and probabilistic fact $0.05 :: \mathrm{stroke}$.
  The rule is decomposed into the probabilistic fact $0.2 :: E$ and the clause
  \[ \mathrm{weakness} \leftarrow \mathrm{stroke}, E.\]
  The induced distribution on the probabilistic fact(s) and error term(s) has four truth value assignments, which specify whether $E$ and/or $\mathrm{stroke}$ are true.
  In this example, the assignment where both are true has probability $0.01$, the assignment where only $E$ is true has probability $0.19$, the assignment where only $\mathrm{stroke}$ is true has probability $0.04$ and the assignment where neither is true has probability $0.76$.
  Weakness holds only where both $E$ and $\mathrm{stroke}$ is true and thus with a probability of $0.01$. 
\end{example}

In addition, ProbLog also allows for disjunctive probabilistic rules of the form
\[ p_1 :: H_1 ; \dots ; p_n:: H_n \leftarrow B_1, \dots, B_n.\]
where $p_1, \dots, p_n$ sum up to at most 1. They encode that, if the body is true, each head $H_i$ is true with probability $p_i$ and the choices of head are mutually exclusive; for details, we refer to  \cite{DeRaedtK15}.  

The programs encoding our generative stroke and TIA models have two parts, one part  covering stroke incidence in the general study population and other reasons for stroke-like symptoms, and the other part covering symptoms due to stroke or TIA.
The independence of the error terms, together with the standard Datalog reading of separate clauses with the same head as a disjunction, allows us to simplify the programs by keeping the auxiliary variables  $S_{\mathrm{stroke}}$ and  $S_{\textrm{other causes}}$ implicit, a mechanism known as ``noisy-or aggregation'' in probabilistic logic programming \cite{DeRaedtK15}. 

The first part deals with the baseline probabilities of the various symptoms in non-stroke-patients and the likelihood for a stroke in the study population as a whole.
We use the data contained in the ``Respondents'' column of Table \ref{tab:simplified_symptoms}, which shows the number of individuals recorded with any of a list of symptoms in a general population of respondents which are not stroke patients during a given two-month interval.
A notable issue is the lack of interaction data between the symptoms.
Hence, the maximum entropy principle implies that  symptoms for other causes then stroke or TIA are to be modelled as independent \cite{Williamson04}.

Another challenge  is that arm and leg weakness are not separated, and neither are facial palsy and speech impairment.
For arm and leg weakness, we deal with the issue by combining the two categories.
This also makes sense in the overall application context, since both of these will be provided by the user directly.
On the other hand, it was important for us to separate facial palsy from speech and language impairment, as facial palsy is eventually to be estimated by a neural classifier.
Hence, the maximum entropy principle implies modelling both symptoms as independent and equally likely outside of stroke patients.
With those assumptions, we can encode the relative frequencies of symptoms in a healthy population in ProbLog facts as in Listing \ref{lst:nullmodel}.  
\begin{lstlisting}[caption = {Model for the occurrence of symptoms within a population of non-patients of stroke or TIA}, label = {lst:nullmodel}, float]
0.078 :: weakness.
0.063 :: sensory.
0.047 :: visual. 
0.005 :: facial_palsy.
0.005 :: speech.
\end{lstlisting}
The likelihood of (the different types of) strokes and TIAs in the study population is also taken from Claus et al.\ \cite{ClausBBLRIW24}, and is modelled by the probabilistic fact $\texttt{0.002 :: stroke\_or\_tia}$ and, for the individual subtypes, the disjunctive rule
\begin{lstlisting}
  (409/900) :: tia; (254/900) :: minor; (237/900) :: major :-
        stroke_or_tia.
\end{lstlisting}

We then move on to model the symptoms of stroke.
Unlike for the reference population of respondents, we do have interaction data between the symptoms of stroke and TIA patients available in Figure \ref{fig:full_symptoms}.
To assess the impact of independence assumptions for diagnostic reasoning in stroke patients, we encode both a simplified maximum entropy model, where the incidence of individual symptoms is taken from Table \ref{tab:simplified_symptoms} and assumed to be independent conditional on stroke or TIA, and a full model, where the entire interaction data from Figure \ref{fig:full_symptoms} is taken into account.  
  
The simplified model encodes the probabilities of the different symptoms conditioned on ``stroke or TIA'' directly, as illustrated in Listing \ref{lst:simplified}.
Note that according to the semantics of ProbLog, the separate clauses with the same body $\texttt{stroke\_or\_tia}$ for each symptom to be caused by stroke render the symptoms conditionally independent given that the body is true, as suggested by entropy maximisation.
\begin{lstlisting}[caption = {Simplified model of stroke symptomatology}, label={lst:simplified}, float]
  weakness :- arm_weakness.
  weakness :- leg_weakness.
  
  0.23 :: facial_palsy :- stroke_or_tia. 
  0.49 :: speech :- stroke_or_tia.
  0.48 :: arm_weakness :- stroke_or_tia.
  0.38 :: leg_weakness :- stroke_or_tia.
  0.23 :: sensory :- stroke_or_tia. 
  0.23 :: visual :- stroke_or_tia.
\end{lstlisting}

Table \ref{tab:simplified_symptoms} also contains more detailed information on the relative frequency of every symptom given the subtype of stroke or TIA. This information also leads to different binary predictions for stroke or TIA, as the symptoms are now no longer assumed to be independent given the binary information on whether stroke or TIA occurred, but only given the precise subtype of stroke or TIA.
Hence, we can capture that portion of correlation that results from the fact that, say, major strokes are more likely to have a combination of different symptoms.      

\begin{lstlisting}[caption = {Model of stroke symptomatology divided by stroke type and TIA}, label={lst:divided}, float]
  stroke_or_tia :- tia.
  stroke_or_tia :- minor.
  stroke_or_tia :- major.
  
  weakness :- arm_weakness.
  weakness :- leg_weakness.
  
  (36/409) :: facial_palsy :- tia. 
  (176/409) :: speech :- tia.
  (96/409) :: arm_weakness :- tia.
  (69/409) :: leg_weakness :- tia. 
  (100/409) :: visual :- tia.
  
  (47/254) :: facial_palsy :- minor. 
  (99/254) :: speech :- minor.
  (126/254) :: arm_weakness :- minor.
  (89/254) :: leg_weakness :- minor.
  (68/254) :: sensory :- minor. 
  (46/254) :: visual :- minor.
  
  (126/237) :: facial_palsy :- major. 
  (167/237) :: speech :- major.
  (211/237) :: arm_weakness :- major.
  (183/237) :: leg_weakness :- major.
  (80/237)  :: sensory :- major. 
  (62/237)  :: visual :- major.
\end{lstlisting}

To create the full model taking account of all the correlations observed by study \cite{ClausBBLRIW24}, we use an auxiliary predicate for every combination tabulated from Figure \ref{fig:full_symptoms} and then record the relative frequency of that combination as well as the symptoms that occur in it. An excerpt is shown in Listing \ref{lst:full}; the full ProbLog program has 54 combinations, including those where none of the included symptoms were recorded:

\begin{lstlisting}[caption = {Snippet of the full model of stroke symptomatology taking co-ocurrence likelihoods into account}, label={lst:full}, float]
  (9/900)::c_1; (11/900)::c_2; (8/900)::c_3; (11/900)::c_4; ...
:- stroke_or_tia.

tia :- c_1.
speech :- c_1.
sensory :- c_1.
minor :- c_2.
sensory :- c_2.
minor :- c_3.
speech :- c_3.
facial_palsy :- c_3.
tia :- c_4.
speech :- c_4.
facial_palsy :- c_4.
...
\end{lstlisting}

\section{The discriminative model}
\label{sec:discriminative}
We wish to integrate neural networks for diagnosing symptoms from visual or auditory data.
We therefore move beyond ProbLog to DeepProbLog \cite{ManhaeveDKDD21}, which incorporates neural components processing sensor data as neural facts, analogous to probabilistic facts.
This requires a model where symptoms can be supplied as facts rather than as evidence. 
Hence, we need to refactor our model from a generative structure, which follows the causal ordering, to a discriminative diagnostic model, in which stroke is at the head of clauses whose body atoms are symptoms.

As the outcome of Section \ref{sec:generative} is completely specified ProbLog models, we can use conditional inference in ProbLog to compute the probabilities of stroke or TIA given various symptom combinations.
We can therefore create a discriminative model by including one probabilistic clause for every fully specified symptom combination.
Listing \ref{lst:discriminative} shows the form of the clauses. 

\begin{lstlisting}[caption = {Example clause from the discriminative model dervied from the model of Listing \ref{lst:divided}}, label={lst:discriminative}, float]
0.41050281 :: stroke_or_tia :-
           weakness,\+facial_palsy,speech,\+sensory,\+visual.
\end{lstlisting}

Regardless of the simplicity of the initial generative causal model, the resulting diagnostic model will always have one rule for every possible combination of symptoms.
To test whether a simpler, more idiomatic model which combines fewer rules could be used, we train the probabilistic inductive logic programming system ProbFOIL 2 \cite{DeRaedtDTVVYW15} using one probabilistic example for each combination as training examples.

ProbFOIL 2 induces a set of probabilistic rules from a set of probabilistic facts.
This setting is known as \emph{learning from probabilistic entailment} \cite{DeRaedtDTVVYW15}, generalising the common setting of learning from entailment known from inductive logic programming. 
\begin{definition}
  Let $\mathcal{P}$ be a set of ground probabilistic facts with respect to a signature $\sigma$ and domain $D$, and let $\mu$ be a probability distribution on $\sigma$ augmented with $D$.
  Then we make the following definitions: The \emph{true positive rate} $\mathrm{TP}(\mu)$ is given as $\sum_{p::\varphi \in \mathcal{P}}{\min(\mu(\varphi),p)}$, the \emph{false positive rate} $\mathrm{FP}(\mu)$ is given as $\sum_{p::\varphi \in \mathcal{P}}{\max(0,\mu(\varphi)-p)}$, the \emph{true negative rate} $\mathrm{TN}(\mu)$ is $\left| \mathcal{P} \right| - \mathrm{FP}(\mu)$ and the  \emph{false negative rate} $\mathrm{FN}(\mu)$ is $\left| \mathcal{P} \right| - \mathrm{TP}(\mu)$ 

  This gives rise to probabilistic versions of the usual derived measures for classification:
  The \emph{(probabilistic) accuracy} is defined as $(\mathrm{TP}(\mu) + \mathrm{TN}(\mu)) / \left| \mathcal{P} \right|$,
  \emph{(probabilistic) precision} is defined as $\mathrm{TP}(\mu)/(TP(\mu) + FP(\mu))$, and \emph{(probabilistic) recall} is defined as  $\mathrm{TP}(\mu)/(TP(\mu) + FN(\mu))$. 
\end{definition}

We make use of ProbFOIL 2 for a form of \emph{theory compression}: given a (complex) probabilistic logic program $\Pi$, we enumerate the conditional probabilities predicted by  $\Pi$ in a set of situations of interest (as ``artificial people'') and we specify them as probabilistic facts.
A snippet of the example file is shown in Listing \ref{lst:examples}.
\begin{lstlisting}[caption={Snippet of the ProbFOIL+ input examples derived from the full model of Listing \ref{lst:full}}, label ={lst:examples}, float]
0.00024264874 :: stroke_or_tia(person1).
0.0035982738 :: stroke_or_tia(person2).
weakness(person2).
0.0029062474 :: stroke_or_tia(person3).
facial_palsy(person3).
0.088949564 :: stroke_or_tia(person4).
weakness(person4).
facial_palsy(person4).
0.062830623 :: stroke_or_tia(person5).
speech(person5).
0.34519212 :: stroke_or_tia(person6).
weakness(person6).
speech(person6).
\end{lstlisting}

From this input, ProbFOIL 2 induces a probabilistic logic program that optimises probabilistic accuracy, balanced with a significance criterion based on the likelihood ratio test.
In this setting and with default parameters, ProbFOIL 2 induced the theory shown in Listing \ref{lst:ProbFOIL}. 
\begin{lstlisting}[caption = {Discriminative model induced by ProbFOIL from the examples of Listing \ref{lst:full}}, label={lst:ProbFOIL}, float]
0.71219084::stroke_or_tia :- speech, facial_palsy.
0.55497685::stroke_or_tia :- speech, weakness.
0.50723151::stroke_or_tia :- sensory, weakness, facial_palsy.
0.3643617::stroke_or_tia :- speech, sensory.
0.74193908::stroke_or_tia :- weakness, visual, sensory, facial_palsy.
0.97247086::stroke_or_tia :- weakness, speech, facial_palsy.
\end{lstlisting}

\section{The neuro-symbolic system}

We integrate a connectionist component acting directly on camera data to identify facial palsy.
Such a component is modelled in DeepProbLog by including a \emph{neural fact} of the form
\[ \mathrm{nn}(m,[X_1, \dots, X_n]) :: r(X_1, \dots, X_n) \]
where $m$ specifies a neural network model (with both architecture and parameters) taking values for the $X_1, \dots, X_n$ as input and providing normalised output between 0 and 1. 
A neural fact can be seen as directly analogous to a probabilistic fact, where, instead of specifying a fixed probability, the program specifies a neural network model.
The output of the neural network model is then used as a probability for this fact. 

In our proof of concept, the classifier is a convolutional neural network (CNN) implemented in Pytorch and trained on some Kaggle data, closely following the architecture described by Dua and Sharma \cite{DuaS25}.
This includes a normalisation layer, four sequential convolutional layers scaling at 16, 32, 64, and 128 filters with max pooling layers applied immediately after every convolutional operation.
Furthermore, 25\% dropout has been applied after every convolutional block to prevent overfitting.
Then, the output is flattened and passed into a 256-unit dense layer with ReLU activation.
We implemented standard $3\times 3$ convolutional kernels and $2\times 2$ pool sizes for the feature extraction layers.
The network was compiled using an Adam optimiser and He/Kaiming weight initialisation.

The neural network parameters were trained on the dataset provided by Elhanashi et al.\ \cite{ElhanashiSZ24}.
This comprises 2,500 negative  and 1,245 posotove examples (i.e. examples with facial palsy), which had been diversified by flipping, rotating, and scaling to match real-world scenarios more closely.

Our system is completed by a simple proof-of-concept \texttt{streamlit} web interface.
Here, the manually collected symptoms and a current facial image are queried from the user.
Then, the discriminative model of Section \ref{sec:discriminative} is dynamically augmented with the user's symptom input
and an image to be processed by the CNN classifier, and the resulting model string is passed to the DeepProbLog engine, which is conveniently available as a Python library.
The system repository, including the models of the preceding sections and associated scripts, can be found at \url{https://github.com/3N61N33R/stroke-detection}.  
\section{Evaluation}
We evaluate the four different probabilistic models for binary prediction of stroke or TIA by comparing their probabilistic accuracy, precision and recall scores as defined by De Raedt et al.\ \cite{DeRaedtDTVVYW15}.
As the ground truth, we use the full model encoding all the combinations of symptoms reported in Figure \ref{fig:full_symptoms}, and we give equal weight to all possible combinations of symptoms.
The results are given in Table \ref{tab:accuracy_metrics}:
\begin{table}
\centering
\begin{tblr}{
    colspec = {llll},
    row{1} = {font=\bfseries}
  }
  Model                         & Accuracy        & Precision  & Recall   \\ \hline
  Simplified  (Listing \ref{lst:simplified})               & 0.919   & 0.830    & 0.989            \\
  Divided (Listing \ref{lst:divided})                   &   0.934   & 0.869      & 0.972           \\ 
  ProbFOIL (Listing \ref{lst:ProbFOIL})               & 0.960 & 0.962 & 0.931 \\
\end{tblr}
\caption{Probabilistic accuracy, precision and recall of the divided, simplified and ProbFOIL-induced models compared to the full model of Listing \ref{lst:full}.}
  \label{tab:accuracy_metrics}
\end{table}

In the DeepProbLog system, we also integrate a neural predicate for detecting facial asymmetry.
The performance of the neural predicate in isolation has been evaluated using an 80-20 train-test split on the available data. It is seen to perform very well, with accuracy 0.9987, precision 0.9958 and perfect recall, as apart from one single misclassified negative example, all samples were classified correctly. 

Overall, the system is designed to perform neuro-symbolic inference from individual data in real time;
to evaluate its runtime, we  ran tests over 100 iterations with varying images and numbers of symptoms.
The results are summarised in Table \ref{tab:runtime_metrics}. 

\begin{table}
  \centering
  \begin{tblr}{
    colspec = {lllll},
    row{1} = {font=\bfseries}
  }
  Component             & Mean (ms) & Median (ms) & Min (ms) & Max (ms) \\ \hline
  CNN Inference         & 46.21     & 46.66       & 32.03        & 56.94        \\
  Model construction    &  5.40     &  5.47       &  3.05        &  8.27        \\
  DeepProbLog inference &101.54     &101.85       & 51.74        &137.88        \\
  Total                 &153.24     &153.16       & 97.70        &201.80         \\
\end{tblr}
\caption{Execution time of CNN inference, construction of the DeepProbLog model string and DeepProbLog inference over 100 iterations with varying images and symptoms}
\label{tab:runtime_metrics}
\end{table}

\section{Discussion, extensions and further work}  
Our architecture showcases the value of DeepProbLog for diagnostic reasoning using neural sensor data, but also shows the complexity that can be incurred for disciplined probabilistic modelling by a design that treats neural predicates as soft facts in the program.

As the neural network is stored as part of the program itself, it can be executed directly in the individual's system.
This mode of deployment avoids privacy issues and issues of data protection as none of the data provided by the individual ever leaves their device.
Probabilities are calculated locally and can then be shown to the user, or, either additionally or alternatively, recommendations for further actions can be displayed that can be  derived  from comparing the calculated probabilities to fixed thresholds.
Alternatively, given sufficient permissions obtained from the user, the program could be made available as a remote service.
DeepProbLog also allows for the gradual evolution of the system, where, as more data is obtained from individuals who have provided full informed consent, the statistical probabilities obtained from the literature in the initial system set-up are successively replaced with parameters learned from available data.
Here, one could opt for either retaining the neural network as-is and only applying parameter learning to the probability annotations, or, if permission is given for the use of individuals' image data for training, DeepProbLog's end-to-end training could be used to train the image-recognising neural network together with the probabilistic parameters.
This flexibility can be a crucial asset for diagnostic systems that can work in a variety of settings, fully functional while running  locally without access to any additional training data while also able to make use of any training data that does happen to be available.

Another dimension of extensibility is the modularity of the system and its capacity to take into account additional categories of input data.
One possible extension is the deployment of additional neural components.
For instance, there has been substantial work on neural systems that detect anomalies of speech, one of the key stroke symptoms and component of the classic FAST test \cite{BanerjeeSJGRRSDG25}.
So, instead of treating speech issues as an item to be supplied by the user, we can simply add a neural network predicting speech issues as a second neural fact to the discriminative model, without any alteration to the remainder of the system.

Similarly, statistical knowledge about additional symptoms can be easily added to the generative model through additional probabilistic rules, with the modularity of the underlying ProbLog program ensuring that this does not impact modelling of the symptoms already accommodated in the current model.
Note, however, that adding one symptom will lead to a discriminative model with twice as many clauses and a completely different set of probabilities, so on this level modularity is a feature of the generative model only.
Of particular practical and also theoretical  interest would be the addition of risk factors such as obesity, known conditions or lifestyle.
Simply adding constraints on the conditional probabilities of stroke given risk factors to the entropy maximisation problem results in a model where symptoms and risk factors are on a single level, all of them rendered independent by conditioning on the disease.
This is unsatisfactory already from a probabilistic perspective, as one would expect the well-known ``explaining-away effect'' to imply that, if we know a stroke victim already has some risk factors, this suffices to accommodate for their elevated stroke risk, thereby rendering other risk factors comparatively less likely \cite{Williamson04}.
Hence, Williamson \cite{Williamson04} argues that the causal distinctions between risk factors and symptoms should be accommodated in the entropy maximisation process, yielding the expected and more satisfactory structure encoded by a Bayesian network on a causal diagram of the form 
\[ \textrm{Risk factors}\longrightarrow\textrm{Illness}\longrightarrow\textrm{Symptoms}. \]
Even so, however, separate ProbLog clauses for each risk factor modelling them as independent causes combined with noisy-or aggregation \cite{DeRaedtK15} do not generally maximise entropy relative to this structure, as noisy-or aggregation encodes an asymmetry between a proposition and its negation, while entropy maximisation knows no such distinction.
Overall, one may argue that in the presence of incomplete statistical information on risk factors, the generative model is less canonical than when reasoning about symptoms only.


Using DeepProbLog for the diagnostic reasoning task raises other interesting issues for model-based neuro-symbolic systems.
The first directly concerns the role of the connectionist components in a DeepProbLog model.
Dealing with raw sensor data is a rather typical use case of connectionist components within a DeepProbLog program.
A very natural role for sensor readings within a probabilistic model is that of an observation.
However, the DeepProbLog architecture supports connectionist components only as neural facts,
necessitating the restructuring of the program from a (potentially simple) causal model to a complex discriminative model.
This suggests an architecture supporting neural predicates as soft evidence (the traditional role for observations in a probabilistic logic program) rather than as neural facts while maintaining as much as possible of the clear semantics and flexible learning and inference provided by (Deep)ProbLog as an enticing prospect for future research.
The relationship between the generative and the corresponding discriminative model is also interesting from a more general viewpoint; we would argue that in diagnostic models such as ours, the generative causal model should be considered as the human-facing representation while the discriminative model may be viewed as more of an implementation artefact.
This has consequences for metrics such as model complexity, where in the maximum entropy model, say, independence assumptions drastically reduce the complexity of the generative model, while the number of parameters of the corresponding discriminative model is exponential in the number of symptoms considered.
So, when transitioning to inducing parameters from data, for instance, the number of parameters to train cannot be easily controlled.

ProbFOIL 2 offers an approach to theory compression that operates directly on the discriminative model itself, rather than on the generative model.
Diagnostic systems from probabilistic information offer a very natural situation for the probabilistic inductive logic programming setting of learning from probabilistic entailment, which seems currently significantly underutilised.   
As  Table \ref{tab:accuracy_metrics} shows, the model learned by ProbFOIL 2 has superior accuracy over the models assuming conditional independence of symptoms, despite having fewer clauses in their discriminative model than those generative models have overall.
However, despite superior accuracy, recall is  poorer for the ProbFOIL 2 model, particularly as many lower probabilities of stroke are rendered deterministically 0.
As in many applications from medical diagnosis, false negatives have generally higher stakes than false positives and even small likelihoods of a serious condition are relevant, this may not be acceptable in practice.    
This imbalance may be a result of accuracy being used as the target metric rather than employing a tunable trade-off metric such as the parametric family of $F_\beta$-scores \cite[p.\ 255]{KroeseBTV20}.
An additional simple enhancement that would enhance the usability of ProbFOIL 2 would be to allow for weighting examples, so that not every case is given equal weight in the accuracy calculation.   
More conceptually, a system that could handle multi-target learning in a setting of learning from probabilistic entailment would allow for theory compression already at the generative level rather than only at the discriminative level.
It would be very interesting to see how those approaches would compare directly.

\begin{credits}
  \subsubsection{\ackname}
  The authors thank Dr Kailin Weitkämper for proof-reading the manuscript.

  \subsubsection{\discintname}
  The authors have no competing interests to declare that are relevant to the content of this article. 
\end{credits}
%
%
%
 \bibliographystyle{splncs04}
 \bibliography{medicalbib,plpbib}

\begin{thebibliography}{10}
\providecommand{\url}[1]{\texttt{#1}}
\providecommand{\urlprefix}{URL }
\providecommand{\doi}[1]{https://doi.org/#1}

\bibitem{BanerjeeSJGRRSDG25}
Banerjee, O., et~al.: Analysis and development of clinically recorded
  dysarthric speech corpus for patients affected with various stroke
  conditions. Neurosci. Inform.  \textbf{5}(2),  100198 (2025).
  \doi{10.1016/j.neuri.2025.100198}

\bibitem{ChenZXGYZLWL21}
Chen, X., Zhao, X., Xu, F., Guo, M., Yang, Y., Zhong, L., Weng, X., Liu, X.: A
  systematic review and meta-analysis comparing fast and befast in acute stroke
  patients. Front. Neurol.  \textbf{12},  765069 (2022)

\bibitem{ClausBBLRIW24}
Claus, J.J., Berghout, B.B., Box, C.V., Licher, S., Roozenbeek, B., Ikram,
  M.K., Wolters, F.J.: Characterizing {TIA} and stroke symptomatology in a
  population-based study: implications for and diagnostic value of fast-based
  public education. BMC Public Health  \textbf{24}, ~3512 (2024)

\bibitem{DeRaedtDTVVYW15}
De~Raedt, L., Dries, A., Thon, I., Van~den Broeck, G., Verbeke, M., Yang, Q.,
  Wooldridge, M.: Inducing probabilistic relational rules from probabilistic
  examples. In: Proc. IJCAI '15. pp. 1835--1842 (2015)

\bibitem{DeRaedtK15}
{De Raedt}, L., Kimmig, A.: Probabilistic (logic) programming concepts. Mach.
  Learn.  \textbf{100}(1),  5--47 (2015). \doi{10.1007/S10994-015-5494-Z}

\bibitem{DuaS25}
Dua, V., Sharma, S.: Detection of stroke-induced facial paralysis using a
  convolutional neural network. Int. J. Softw. Hardware Res. Eng.
  \textbf{13}(6),  8--15 (2025). \doi{10.26821/IJSHRE.13.06.2025.130601}

\bibitem{ElhanashiSZ24}
Elhanashi, A., Saponara, S., Zheng, Q.: Annotation facial images for stroke
  classification acute vs non acute. In: 2024 Sixth International Conference on
  Intelligent Computing in Data Sciences (ICDS) (2024).
  \doi{10.1109/ICDS62089.2024.10756361},
  \url{https://www.kaggle.com/datasets/abdussalamelhanashy/annotated-facial-images-for-stroke-classification}

\bibitem{FierensVanDenBRSGTJD2015}
Fierens, D., Van Den~Broeck, G., Renkens, J., Shterionov, D., Gutmann, B.,
  Thon, I., Janssens, G., {De Raedt}, L.: Inference and learning in
  probabilistic logic programs using weighted boolean formulas. Theory Pract.
  Log. Program.  \textbf{15},  358–401 (2015)

\bibitem{Hofman15}
Hofman, A., et~al.: The {Rotterdam Study}: 2016 objectives and design update.
  Eur. J. Epiemiol.  \textbf{30}(8),  661--708 (2015)

\bibitem{KroeseBTV20}
Kroese, D.P., Botev, Z.I., Taimre, T., Vaisman, R.: Data Science and Machine
  Learning. Routledge (2020)

\bibitem{ManhaeveDKDD21}
Manhaeve, R., Dumančić, S., Kimmig, A., Demeester, T., {De Raedt}, L.: Neural
  probabilistic logic programming in {DeepProbLog}. Artif. Intell.
  \textbf{298},  103504 (2021)

\bibitem{ParisV90}
Paris, J.B., Vencovsk{\'{a}}, A.: A note on the inevitability of maximum
  entropy. Int. J. Approx. Reason.  \textbf{4}(3),  183--223 (1990).
  \doi{10.1016/0888-613X(90)90020-3}

\bibitem{ParisV97}
Paris, J.B., Vencovsk{\'{a}}, A.: In defense of the maximum entropy inference
  process. Int. J. Approx. Reason.  \textbf{17}(1),  77--103 (1997).
  \doi{10.1016/S0888-613X(97)00014-5}

\bibitem{ShoreJ80}
Shore, J.E., Johnson, R.W.: Axiomatic derivation of the principle of maximum
  entropy and the principle of minimum cross-entropy. {IEEE} Trans. Inf. Theory
   \textbf{26}(1),  26--37 (1980). \doi{10.1109/TIT.1980.1056144}

\bibitem{Williamson04}
Williamson, J.: Bayesian Nets and Causality: Philosophical and Computational
  Foundations. Oxford University Press (2004)

\end{thebibliography}
%





\end{document}